\documentclass[11pt]{article}
\usepackage{nlp4call2021}
\usepackage{times}
\usepackage{url}
\usepackage{latexsym}

\usepackage{rotating}
\usepackage{tabularx}
\usepackage{graphicx}
\usepackage{booktabs} 
\usepackage{caption}
\usepackage[T1]{fontenc}
\usepackage[utf8]{inputenc}
\usepackage{todonotes}

\usepackage{appendix}

\newcommand{\DEVELOPMENT}{1} 
\usepackage{ifthen}
\ifthenelse{\DEVELOPMENT = 1}{
	\newcommand{\ev}[1]{\textcolor{cyan}{\textbf{EV:} #1}}	
	\newcommand{\sam}[1]{\textcolor{magenta}{\textbf{SAM:} #1}}
    \newcommand{\jk}[1]{\textcolor{blue}{\textbf{JK:} #1}}
}{
	\newcommand{\ev}[1]{}		
	\newcommand{\sam}[1]{}
    \newcommand{\jk}[1]{}
}

\newcommand{\ANON}{0}

\ifthenelse{\ANON = 1}{
\newcommand{\anonurld}[1]{\textsc{Anonymized URL}}
\newcommand{\anon}[1]{\textsc{Anonymized}}

\let\url=\anonurld
\newcommand{\anoncitep}[1]{\textsc{(Anonymized citation)}}
\newcommand{\anoncitet}[1]{\textsc{Anonymized citation}}
}{
\newcommand{\anon}[1]{#1}

\newcommand{\anoncitep}[1]{\citep{#1}}
\newcommand{\anoncitet}[1]{\citet{#1}}
}

\aclfinalcopy 

\title{DaLAJ - a dataset for linguistic acceptability judgments for Swedish:\\ Format, baseline, sharing}

\author{Elena Volodina\textsuperscript{1}, Yousuf Ali Mohammed\textsuperscript{1}, Julia Klezl\textsuperscript{2}\\
 University of Gothenburg, Sweden \\
  {\tt \textsuperscript{1}name.surname1.surname2@svenska.gu.se}\\
   {\tt \textsuperscript{2}gusklezju@student.gu.se}}

\date{}

\begin{document}
\maketitle
\begin{abstract}
We present DaLAJ 1.0, a \textbf{Da}taset for \textbf{L}inguistic \textbf{A}cceptability \textbf{J}udgments for Swedish, comprising 9\,596 sentences in its first version; and the initial experiment using it for the binary classification task.\ DaLAJ is based on the SweLL second language learner data \citep{volodina2019swell}, consisting of essays at different levels of proficiency.\ To make sure the dataset can be freely available despite the GDPR regulations, we have sentence-scrambled learner essays and removed part of the metadata about learners, keeping for each sentence only information about the mother tongue and the level of the course where the essay has been written.\  We use the normalized version of learner language  as the basis for the DaLAJ sentences, and keep only one error per sentence.\ We repeat the same sentence for each individual correction tag used in the sentence.\ For DaLAJ 1.0 we have used four error categories (out of 35 available in SweLL), all connected to lexical or word-building choices.\ Our baseline results for the binary classification show an accuracy of 58\% for DaLAJ 1.0 using BERT embeddings. The dataset is included in the SwedishGlue (Swe. SuperLim) benchmark. 
Below, we describe the format of the dataset, first experiments, our insights and the motivation for the chosen approach to data sharing.

\let\thefootnote\relax\footnotetext{This work is licensed under a Creative Commons Attribution 4.0 International Licence. Licence details: http://creativecommons.org/licenses/by/4.0/.}

\end{abstract}


\section{Introduction}
\label{sec:related}


\begin{table*}[!hbt]
    \centering
    \begin{tabular}{llrrrr}
    \toprule
         \bf Categories & \bf Explanation & \bf A-lev & \bf B-lev & \bf C-lev & \bf Total \\ \midrule
         O-Comp & Problem with compounding & 252 & 62 & 232 & 546 \\
        L-Der & Word formation problem (derivation or compounding) & 193 & 124 & 404 & 721 \\
        L-FL & Non-Swedish word corrected to Swedish word & 46 & 17 & 26 & 89 \\
        L-W & Wrong word or phrase & 1157 & 562 & 1723 & 3442 \\
         {\bf Total} & & {\bf 1648} & {\bf 765} & {\bf 2385} & {\bf 4798} \\
         \bottomrule
    \end{tabular}
    \caption{Dataset overview, with number of sentences per correction tag, level and in total}
    \label{tab:datasetStats}
\end{table*}

Grammatical and linguistic acceptability is an extensive area of research that has been studied for generations by theoretical linguists \cite[e.g.][]{chomsky1957syntactic}, and lately by cognitive and computational linguists \cite[e.g.][]{keller2000gradience,lau2020furiously, warstadt2019neural}. 
Acceptability of sentences is defined as "the extent to which a sentence is permissible or acceptable to native speakers of the language." \cite[][p.1618]{lau2015unsupervised}, and there have been different approaches to studying it.\ 
Most work views acceptability as a binary phenomenon:\ the sentence is either acceptable/ grammatical or not \cite[e.g.][]{warstadt2019neural}.\ \citet{lau2014measuring} show that the phenomenon is in fact gradient and is dependent on a larger context than just one sentence.\ 
While most experiments are theoretically-driven, the practical value of this research has been also underlined, especially with respect to language learning and error detection \cite{wagner2009judging,daudaravicius2016report}.

 






Datasets for acceptability judgments require linguistic samples that are unacceptable, which requires a source of so-called negative examples. Previously, such samples have been either manually constructed, artificially generated through machine translation \cite{lau2020furiously}, prepared by automatically distorting acceptable samples e.g. by deleting or inserting words or inflections \cite{wagner2009judging} or collected from theoretical linguistics books \cite{warstadt2019neural}.\ Using samples produced by language learners has not been mentioned in connection to acceptability and grammaticality studies. However, there are obvious benefits of getting authentic errors that automatic systems may meet in real-life.  
Another benefit of reusing samples from learner corpora is that they often contain not only corrections, but also labels describing the corrections. The major benefit, though, is that (un)acceptability judgments come from experts, 
i.e. teachers, assessors or trained assistants, and are therefore reliable. 


\section{Dataset description}
\label{sec:dataset}

\begin{table}
    \centering
    \begin{tabular}{llr}
    \toprule
         \bf Approximate level & \bf Nr essays & \bf Nr labels \\ \midrule
         A:Beginner & 289 & 11\,180 \\
         B:Intermediate & 45 & 5\,119 \\
         C:Advanced & 168 & 12\,986 \\
         {\bf Total} & {\bf 502} & {\bf 29\,285} \\
         \bottomrule
    \end{tabular}
    \caption{Statistics over the SweLL data}
    \label{tab:swell}
\end{table}

We use the error-annotated learner corpus SweLL \cite{volodina2019swell} as a source of "unacceptable" sentences and select sentences containing corrections of the type that is of relevance to the SwedishGlue benchmark\footnote{SwedishGlue is a collection of  datasets for training and/or evaluating language models for a range of Natural Language Understanding (NLU) tasks.} \cite{adesam2020swedishglue}. In the current version, four {\it lexical error types} are included into the DaLAJ dataset (see Section \ref{sec:err_type}). The resulting dataset contains 4\,798 sentence pairs (correct-incorrect), where the two sentences in each sentence pair are identical to each other except for one error. In total, DaLAJ 1.0 contains 9\,596 sentences (which is a sum of unacceptable sentences and their corrected "twin" sentences). To compare, \citet{lau2014measuring} use a dataset of 2\,500 sentences and \citet{warstadt2019neural} have about 10\,700 sentences for a similar task.\ We have a possibility to extend the DaLAJ dataset by other correction types (spelling, morphological or syntactical) in future versions.\ 
The full SweLL dataset contains 29\,285 correction tags, of which 25\,878 may become relevant for the current task (omitting punctuation, consequence and unintelligibility correction tags).

\begin{figure}
    \centering
    \includegraphics[angle=90,origin=c,width=4.6cm]{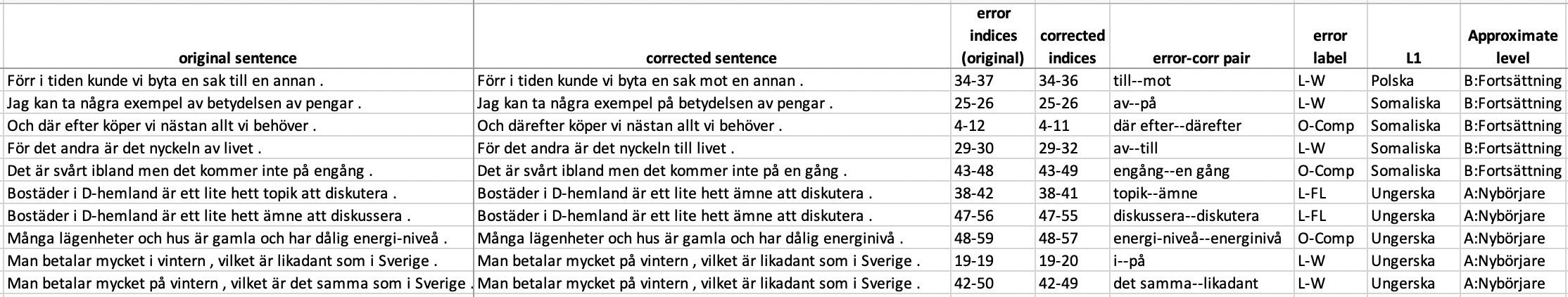}
    \caption{An excerpt from the dataset}
    \label{fig:dataExcerpt}
\end{figure}

\subsection{The source corpus}
The SweLL data \cite{volodina2019swell} has been collected over four years (2017-2020) from adult learners of Swedish from formal educational settings, such as courses and tests. The collection contains about 680 
pseudonymized essays in total, with 502 of those manually normalized (i.e. rewritten to standard Swedish) and annotated for the nature of the correction (aka error annotation). Table \ref{tab:swell} shows the statistics over SweLL in number of essays and correction tags per level. 
Levels of the sentences correspond to the level of the course that learners were taking when they wrote essays. The essays represent several levels, namely:

{\small {\tt A}} - beginner level 

{\small {\tt B}} - intermediate level 

{\small {\tt C}} - advanced level 

The data is saved in two versions: the original and the normalized, with correction labels assigned to the links between the two versions. The 502 corr-annotated essays contain 29\,285 corrections distributed over 35 correction tags, as listed in Appendix A. 

\subsection{Selection of (un)grammatical sentences}
\label{sec:err_type}

The linguistic acceptability task in the SwedishGlue is described as a natural language understanding (NLU) task conceptualized as binary 
judgments 
from a perspective relevant for research on language learning, language planning etc. \cite{adesam2020swedishglue}. Semantic aspects of the sentence are the main focus of this task. This deviates from the type of language included into the CoLA dataset available through GLUE \cite{warstadt2019neural}, where also morphological and syntactic violations are included. In DaLAJ 1.0, we have selected four correction types from the SweLL corpus that would maximally correspond to the need of semantic interpretation of the context, namely {\small {\tt L-W, L-Der, L-FL, O-Comp}} \cite{rudebeck2020correction}, described below.

\paragraph{L-W: Wrong word or phrase.} The {\small {\tt L-W}} tag represents the correction category \textit{wrong word or phrase}. It is used when a word or phrase in the original text has been replaced by another word or phrase in the normalized version. It is placed on units which are exchanged rather than corrected. For example,

\noindent {\small\verb|Alla blir *busiga med sociala medier →|}

\noindent {\small\verb|Alla blir upptagna med sociala medier|}

\noindent which may be verbatim translated as  

\noindent {\small\verb|Everyone is *naughty with social media →|}

\noindent {\small\verb|Everyone is busy with social media|}

\noindent Note the Engligh influence on the use of the word {\small\verb|*busiga|} to convey the meaning that someone is {\small\verb|*busy|} (Swe {\small\verb|upptagen|}), the Swedish word {\small\verb|busig|} meaning {\small\verb|mischievous, naughty|}.

\paragraph{L-Der: Word formation.} The {\small {\tt L-Der}} tag represents the correction category {\it deviant word formation}. It is used for corrections of the internal morphological structure of word stems, both with regard to compounding and to derivation.

The {\small {\tt L-Der}} tag is exclusively used for links between one-word units (not necessarily one-token units, since a word may mistakenly be written as two tokens), where the normalized word has kept at least one root morpheme from the original word, but where another morpheme has been removed, added, exchanged or had its form altered. For example,

\noindent {\small\verb|De är *stressiga på grund av studier →|}

\noindent {\small\verb|De är stressade på grund av studier|}

\noindent which may be translated as  

\noindent {\small\verb|They are *stressy because of the studies →|}

\noindent {\small\verb|They are stressed because of the studies|}

\noindent Note that {\small\verb|*stressiga|} uses an existing derivation affix {\small\verb|-ig(a)|}, which is wrong in this context, instead of the correct suffix {\small\verb|-ade, stressade|}.

\paragraph{L-FL: Foreign word corrected to Swedish.} The {\small {\tt L-FL}} tag is used for {\it words from a foreign (non-Swedish) language} which have been corrected to a Swedish word. It may also be applied to words which have certain non-Swedish traits due to influence from a foreign language. For example,

\noindent {\small\verb|Jag och min *family → |}

\noindent {\small\verb|Jag och min familj|}

\noindent  {\small\verb|English: I and my family|}

\paragraph{O-Comp: Spaces and hyphens between words.} The {\small {\tt O-Comp}} tag is used for corrections which involve the removal of a space between two words which have been interpreted as making up a compound in the normalized text version, or, more rarely, the adding of a space between two words. It may also be used for corrections regarding the use of hyphens in {\it compounds}. Some examples,

\noindent {\small\verb|Jag kände mig *jätte *konstig →|}

\noindent {\small\verb|Jag kände mig jättekonstig|}

\noindent {\small\verb|English: I felt very strange |}


\noindent Distribution of the correction tags in the DaLAJ 1.0 dataset is shown in Table \ref{tab:datasetStats}. 

\subsection{Data format}

The task of linguistic acceptability judgments is traditionally performed on the {\it sentence level}, where each sentence includes {\it maximum one deviation}. In real life learner-written sentences may contain several errors, but it has been shown that training algorithms on samples with focus on one error only produces better results than when mixing several errors in one sentence; extending the context to a paragraph  may further improve the results \cite{katinskaia2021assessing}.\ Paragraphs in learner data, however, are not predictable or well defined, and on several occasions in the SweLL data entire essays consist of one paragraph only. Including in the DaLAJ dataset full paragraphs, in certain cases equivalent to full essays, entails risks of revealing author identities through indications of author-related events or other identifiers despite our meticulous work on pseudonymization of essays \cite{volodina2020towards,megyesi2018learner}. We assess, therefore, that we have no possibility to include paragraphs into the dataset due to the restrictions imposed by the GDPR, so we follow the generally accepted standard of single sentences with single deviations.

For each correction label used in the corpus data, we take the corrected target sentence and preserve only one erroneous segment in it to make it "unacceptable". This means that the same sentence can be repeated several times in the dataset, with different segments/deviations being in the focus. Positive samples are represented by the corrected sentences. We have data in a tab separated file format, with eight columns, namely:

\noindent 1. 
Original (i.e.\ unacceptable) sentence, e.g.\ {\small {\tt Men pengarna är inte *alls}} (Eng.\ But money is not *at all)

\noindent 2. 
Corrected sentence, e.g. {\small {\tt Men pengarna är inte allt}} (Eng. But money is not everything)
   
\noindent 3. 
Error string indices, e.g. {\small {\tt 21-24}}

\noindent 4. 
Correct string indices, e.g. {\small {\tt 21-24}}

\noindent 5. 
Error-correction pair, e.g. {\small {\tt alls-allt}}

\noindent 6. 
Error label, e.g. {\small {\tt L-W}}

\noindent 7. 
Mother tongue(s) (L1), e.g. {\small {\tt Somali}}
    
\noindent 8. 
Approximate level, e.g. {\small {\tt B:Intermediate}}
\\

Figure \ref{fig:dataExcerpt} shows an excerpt from the dataset. Note that some of the sentences in the "Corrected sentence" column are repeated more than once. The corresponding original sentences contain a new error focus each time. The dataset is (by default) balanced with respect to the number of correct and incorrect samples, however, correct samples contain a number of duplicates which should be complimented by a corresponding number of unique correct sentences, which is something we will add in the next release of the dataset. The dataset is not equally balanced as far as number of sentences per level or per correction code are concerned, which is a more challenging problem.

CoLA dataset authors have explicitly tested that the vocabulary used in their dataset belongs to the 100\,000 most frequent words in the language \cite{warstadt2019neural}. In the case of DaLAJ, we have not  done any such investigation since we believe that the vocabulary used by second language learners cannot be so advanced as to be outside the 100K most frequent words.  


The DaLAJ 1.0 dataset is freely available at the SwedishGlue webpage\footnote{https://spraakbanken.gu.se/en/resources/swedishglue}.






\section{Experiments}
\label{sec:experiment}
To set a baseline for linguistic acceptability classification on this dataset, we trained and evaluated four neural network models, using two different types of word embeddings and two different versions of the training and validation sets. All models are sentence-level classifiers based on a bidirectional LSTM layer and a linear output layer. 

\subsection{Word embeddings}
We used two different types of pretrained word embeddings. With smaller datasets like this one, pretrained embeddings have the advantage of being trained on much larger corpora, which generally makes them more informative. The disadvantage here 
is that the domains the embeddings are trained on are quite different from the domain we used them in. 

For two of the models, 300-dimensional Swedish FastText embeddings were used. They were trained on Common Crawl and Wikipedia data, using the CBOW method, character 5-grams, and a window size of 5 \citep{grave2018learning} and contain a vocabulary of 1\,999\,998 words. 

For the other two models, we used contextualized embeddings extracted with the Swedish BERT model\footnote{https://huggingface.co/KB/bert-base-swedish-cased} trained by the National Library of Sweden on a variety of sources including Wikipedia and internet forums as well as books, newspapers, and government publications. It has a vocabulary size of about 50\,000. The BERT embeddings are the sum of the hidden states of the last four encoder layers for every word. This results in 768-dimensional word embeddings. Since these embeddings are context-dependent, the expectation was that they adapt better to the domain of learner language and give better results. 

\subsection{Training, validation, and test data}
We conducted two experiments with each embedding. In Experiment 1, we used the DaLAJ 1.0 dataset for training, validation and testing whereas in Experiment 2 we used the SweLL dataset with 35 error types for training and validation and the same DaLAJ 1.0 test set for testing. As mentioned in Section \ref{sec:dataset}, almost 26\,000 incorrect sentences are available from the SweLL dataset when all error tags are included. We used a split of 80\%, 10\% and 10\% for training, validation and testing in both experiments. 


In both datasets, we used sentences with a maximum length of 50 tokens and below. We excluded the sentences with error types that occur less than 100 times in our dataset as well as duplicate incorrect sentences.
We have a unique vocabulary of 7\,718 for the DaLAJ 1.0 dataset and 12\,325 for the SweLL dataset including the punctuation. Table \ref{tab:train_test_val_data} shows the number of sentences in the train, val, and test datasets after preprocessing.

\begin{table}
\centering
\begin{tabular}{lrrrr}
\toprule 
\bf Experiment  & \bf Train & \bf Val & \bf Test & \bf Vocab \\
\midrule
1 (DaLAJ 1.0)  & 6\,870   & 892      & 952   &  7\,718 \\
2 (SweLL) & 32\,762   & 4\,246      & 952   &  12\,325 \\
\bottomrule
\end{tabular}
\caption{Number of sentences for training, validation, and testing in both experiments}
\label{tab:train_test_val_data}
\end{table}
Due to these filtering steps, {\small {\tt L-FL}} errors were not included in the final datasets, so we only report detailed results for the remaining three categories.

\subsection{Model details}
The machine learning models were trained and evaluated in PyTorch.\footnote{https://pytorch.org}\ We used bidirectional LSTM and linear layers in both the experiments as the input layers in the models are sequential. A softmax function was used to convert the output predictions into probabilities and an argmax function to generate the labels from these probabilites. Below is a list of parameters we used. 

\noindent • Batch size: 32

\noindent • Learning rate:

– \textit{1e$^{-4}$} (FastText embeddings)

– \textit{5e$^{-5}$} (Swedish BERT embeddings)

\noindent • Optimizer: Adam

\noindent • Loss function: CrossEntropyLoss

\noindent • Epochs: 75, Earlystopping: 15 epochs on \textit{validation\_accuracy}


In the models with FastText embeddings we used only the words that occur more than 3 times in our dataset.\ The remaining words were embedded as UNK (\textit{unknown}) in both the training and evaluation process. For the BERT embedding models we used the whole vocabulary set without any cutoff.\ All models were trained for a maximum of 75 epochs with early stopping after 15 epochs without improvements on validation accuracy. The high threshold for early stopping was chosen as the validation accuracy for the models using BERT embeddings were gradually increasing after each epoch. 

\section{Results}
\label{sec:results}

\begin{table*}[!hbt]
\centering
\begin{tabular}{llllll}
\toprule 
\bf Experiment & \bf Embeddings  & \bf Accuracy & \bf Precision & \bf Recall & \bf F1 Score\\ 
\midrule
1 (DaLAJ 1.0) & FastText & 0.540   & 0.55      & 0.45   & 0.49\\ 
2 (SweLL) & FastText   & 0.542   & 0.55      & 0.43  & 0.49 \\ 
1 (DaLAJ 1.0) & BERT     & 0.566   & 0.56 & 0.58  & 0.57\\ 
2 (SweLL) & \bf BERT & \bf 0.583 & \bf 0.57 & \bf 0.69 & \bf 0.62\\ 
\bottomrule
\end{tabular}
\caption{Evaluation results for the DaLAJ test dataset}
\label{tab:models_performance}
\end{table*} 

There are two aspects of the results we are interested in. First, we evaluate the average accuracy, precision, recall, and F1-score of the four models to get an overview of their overall performance. Second, we look at the accuracy per error type to determine which of them are easier or harder 
to learn for the models. 

The results in Table \ref{tab:models_performance} show a clear pattern regarding the first aspect. BERT embeddings are more successful than FastText embeddings in all four metrics, with the biggest margin being between the best and worst model's recall. Training on more data (including additional error types) helps improve the results in all variants, but the improvement is more significant in the models using BERT. Generally, the difference between the FastText and BERT models is bigger than the difference between the models with more or less training and validation data.

\begin{table}[h]
\centering
\begin{tabular}{llrr}
\toprule
\bf Error tag  & \bf Acc. & \bf \# train sent & \bf \# test sent \\ 
\midrule
O-Comp & 0.563 &  474  & 80     \\ 
L-Der  & 0.625 &  1074  & 184    \\ 
L-W    & 0.572 &  5198 & 670    \\ 
\bottomrule
\end{tabular}
\caption{Performance by group for the best model (BERT 2, trained and validated on all errors, tested on DaLAJ)}
\label{tab:accuracy_groups}
\end{table}

As Table \ref{tab:accuracy_groups} shows, there is quite some variation between the error types in the best-performing model. {\small {\tt L-Der}} errors have the highest accuracy, followed by {\small {\tt L-W}}, and {\small {\tt O-Comp}}. Interestingly, the performance differences do not correspond to the number of samples for each correction type. The fact that the best-represented type, {\small {\tt L-W}}, has an accuracy far below the best-performing type, {\small {\tt L-Der}}, indicates that it is inherently more difficult to learn than the others. The reason for this could be that such word choice errors are sometimes either matters of taste or style rather than being clearly incorrect (see [1] below); depend on the context, which is not given in this dataset and task [2]; or deal with use of wrong prepositions, which seems to be very challenging to capture by the models [3].


\noindent {\texttt{\small [1] (Swe) Men den ändras jätteofta för}} 

\hspace{\parindent} {\texttt{\small att *personer vill ha mer.}} 

\hspace{\parindent} {\texttt{\small [*personer → människor]}}

\hspace{\parindent} {\texttt{\small (Eng) But they change very often  }}

\hspace{\parindent} {\texttt{\small since *persons want to have more.}} 

\hspace{\parindent} {\texttt{\small [*persons → people]}}


\noindent {\small\verb|[2] (Swe) Ni får inte.  [får → måste]|}

\hspace{\parindent} {\small\verb|(Eng) You may not.  [may → must]|}


\noindent {\small\verb|[3] (Swe) Om ni vill åka dit måste ni|}

\hspace{\parindent} {\texttt{\small ta båt *mot D-stad. [*mot → till]}} 


\hspace{\parindent} {\small\verb|(Eng) If you want to go there you  |}

\hspace{\parindent} {\texttt{\small must take a boat *towards D-city.}} 

\hspace{\parindent} {\texttt{\small [*towards → to]}}

\section{Discussion}
\label{sec:discussion}

It is difficult to compare our results to previous work done on other datasets (in English) since there is a wide range in performance levels. For example, \citet{warstadt2019neural} reached an accuracy of up to 73\% out-of-domain or 77\% in-domain on their CoLA data, which includes not only semantic, but also syntactical and morphological violations.\ Teams participating in the AESW 2016 shared task reached F-scores between 46\% and 62\% (on the binary task) on data that contained grammatical errors as well as stylistic deviations \cite{daudaravicius2016report}.\ As demonstrated by our results, some error types are significantly easier to model than others, and therefore models trained and evaluated on different sets of error types are not directly comparable and their results should be taken as indicative ones. 

Our analysis has suggested, that the DaLAJ 1.0 dataset needs to be cleaned in several ways. First, the SweLL corpus contains a number of essays where learners add reference lists by the end of essays. Naturally, punctuation in reference lists is non-standard, among others not always containing full stop which sabbotages sentence segmentation. Besides, references are syntactically elliptical and do not fit into the standard language that the embeddings are trained on.\ A part of the errors made by the best-performing model have been observed in sentences representing references.\ We would need to clean the dataset of all such sentences to ensure more objective training and testing.

Second, some sentences where our predictions were erroneous, contain "hanging" titles or e-mail headers. Those hanging elements have not been separated by a full stop in the original essays, and have been prefixed to the next following sentence, making the resulting syntactic structure erroneous despite absent error tags, which of course triggers our models to predict an error, e.g. 
{\texttt{\small (Swe) En B-institution-entusiast Hej Segerstad kommun !}} 
> {\texttt{\small (Eng) A B-institute-enthusiast Hi Segerstad municipality !}}

Third, we have observed that each time a sentence contains a pseudonymized element of the form {\texttt{\small A-city, B-street, C-element}}, etc., it triggers the model to predict an error. We plan to use a simple mapping strategy to replace all patterns for, e.g. cities, with the same city name, etc. 

Yet another observed weakness of the DaLAJ 1.0 dataset, is that the positive sentences are repetitive. Since the models need to be trained on unique samples, we plan to exchange the non-unique ones with other sentences. Luckily, positive samples are easier to find than negative ones. We will use a corpus of L2 coursebooks graded for levels of proficiency, COCTAILL \cite{volodina2014you}, to replace duplicate sentences with the ones of equivalent level, and as far as possible, having similar linguistic features and length.  

There are a number of other patterns in the dataset that seem to cause systematic erroneous predictions, e.g. unintelligibility (X) labels, which we will look into and work around as long as it is possible 
in DaLAJ 1.1 version.

Finally, there is an important difference between the type of sentences used in CoLA and DaLAJ datasets. CoLA sentences are constructed manually for linguistic course books exemplifying various theoretically important linguistic features, and do not require wider context to interpret; whereas DaLAJ sentences are torn out of their natural context, and contain anaphoric references and elliptical structures. However, the applied value of training (machine learning) algorithms on DaLAJ sentences is higher than CoLA sentences (as we imagine that) since such models can be used in language learning context for writing support. 

\section{Reflections on access to learner data}
\label{sec:considerations}

Datasets and corpora collected from (second) language learners contain private information represented both on the metadata level and - depending on the topic - in the texts. Presence of personal information makes those datasets non-trivial to share with the public in a FAIR\footnote{FAIR: Findable, Accessible, Interoperable, Reusable \cite{wilkinson2016fair}} way  \cite{frey2020creating,volodina2020towards}, to say nothing of a potential to use such data for {\it shared tasks}. This is rather unfortunate since collection and preparation of such corpora is an extremely time-consuming and expensive process. Language learner datasets can seldom boast big sizes appropriate for training data-greedy machine learning algorithms, and could therefore benefit from aggregating data from several sources - provided they are accessible. Access to such data, besides, ensures transparency of the research and stimulates its fast development \cite{macwhinney2017shared,marsden2018data}. 

As data owners, we have to face two contradictory forces: one requiring open sharing, and the other preventing it. Among advocates for sharing data openly we see 

\noindent • national and international funding agencies, e.g. Swedish Research Council\footnote{https://www.vr.se/english/mandates/open-science/open-access-to-research-data.html} or European Commission\footnote{https://ec.europa.eu/info/research-and-innovation/strategy/goals-research-and-innovation-policy/open-science/open-access\_en}, requiring guarantees from grant holders that any produced data will be made available for other researchers, 

\noindent • national and international infrastructures, e.g. Clarin\footnote{https://www.clarin.eu/} or SLABank,\footnote{https://slabank.talkbank.org/} and 

\noindent • updated journal policies (e.g. The Modern Language Journal).\footnote{https://onlinelibrary.wiley.com/journal/15404781} 

On the more restrictive side, we have national Ethical Review Authorities\footnote{https://www.government.se/government-agencies/the-swedish-ethics-review-authority-etikprovningsmyndigheten/} and the  General Data Protection Regulation,  GDPR \cite{gdpr2018}, described shortly below. 

\noindent \textbf{The Swedish Ethical Review Authority} currently requires that we keep the original data (e.g.\ hand-written/ non-transcribed/ non-pseudonymized essays) for ten years after the project end so that researchers, who may question the trustworthiness of the original data handling, can require access to the original data for inspection. This means that the data owners need to keep mappings between learner names and their corpus IDs to make it possible to link de-identified and pseudonymized essays to their 
original 
versions. 

\noindent \textbf{General Data Protection Regulation} sets certain limitations on the data where personal data occurs, among others: 
 
\noindent • learner identities should be protected, e.g. pseudonymized or de-identified; 

\noindent • data need to be removed if any of the data providers (=learners) requests that; 

\noindent • users that are granted access to the data should have affiliation inside Europe; 
and 

\noindent • questions that users can work with are limited to the ones stated in the consent forms, in the case of SweLL encompassing research on and didactic applications for language learning. 

To meet these requirements, data owners need to administer data access through an application form, where applicants have to be asked about their geographical location and research questions, and need to be informed about the limitations of spreading data to unauthorized users, etc. Users outside Europe can file an application to the university lawyers who have to consider them on a case-to-case basis. The GDPR applies to the data as long as a mapping of learner names with their corpus IDs (as required by the Ethical Review Authorities) is not destroyed. At a certain point of time (currently 10 years) the mapping key will be destroyed and the data will no longer be under the GDPR protection. 

In both cases, a 10-year quarantine is obligatory. The restrictions above do not seem to hamper most of the potential EU-based researchers from getting access to the data in its entirety, especially researchers working with qualitative analysis of the data inside a limited project group, e.g. Second Language Acquisition researchers or researchers on language assessment. However, when it comes to the NLP field, the most effective way to stimulate research is to organize {\it shared tasks}  or provide access to testing and evaluation datasets without any extra administration, as it is, for example, done in the GLUE\footnote{https://gluebenchmark.com/} and SuperGLUE\footnote{https://super.gluebenchmark.com/} benchmarks \cite{wang2018glue,wang2019superglue}. 

From the above it follows that data owners need to keep a promise to the funding agencies to make the data open, and at the same time, to follow the legislation and keep the data locked within Europe and only for research questions dealing with language learning. 
Being representatives of a “trapped researcher” group, we have been considering how to make learner data available for a wider audience.\ 
For a range of NLP tasks we suggest, thus, sharing L2 data in a sentence scrambled way with limited amount of socio-demographic metadata, for example for error detection \& correction tasks.\ 
The DaLAJ dataset is a proof-of-concept attempt in this direction.

Ultimately, the education NLP community working with L2 datasets would win by setting up a benchmark with available (multilingual) datasets in the same way as GLUE benchmark is doing for Natural Language Understanding (NLU) tasks.

\section{Concluding remarks}
\label{sec:conclusions}

We have presented a new dataset for Swedish which can be used for a variety of tasks in Natural Language Processing (NLP) or Second Language Acquisition (SLA) contexts. We see our contributions both with regards to the dataset, baseline and evaluation metrics, as well as with suggesting a format for L2 datasets that may allow sharing learner data more openly 
in the GDPR age.

The described experiment has become a validity test of the proposed data format and a way to identify its weaknesses and strengths. We see multiple advantages to use the proposed format for L2 data. Apart from a potential to share the data with wider community of researchers, it also (1) helps expand the data (each original sentence potentially generating several sentences) and (2) helps focus on one error only, facilitating fine-grained analysis of model performance as well as human evaluation of model predictions.









In the near future, we will test binary linguistic acceptability classification on the full SweLL dataset (all error tags), per error category and level. We plan to correlate the classification results with correction categories, levels and L1s. Further, we plan to apply models, trained on DaLAJ, to real learner data containing multiple errors per sentence, to assess the effect of data manipulation (i.e. original essays > DaLAJ format) on algorithm training. Proofreading the dataset and addressing identified weaknesses and errors is another direction for the future work. 

In some more distant future we would like to organize shared tasks using DaLAJ.
Apart from binary classification for linguistic acceptability judgments, 
we see a potential of using DaLAJ dataset (in extended version to cover the full correction tagset) for a range of other tasks, including:


\noindent • error detection (identification of error location) 

\noindent • error classification (labeling for error type) 

\noindent • error correction (generating correction suggestions) 

\noindent • first language identification (given samples written by learners, to identify their mother tongues)

\noindent • classification of sentences by the level of proficiency of its writers, and other potential tasks.




\section*{Acknowledgments}
This work has been supported by \emph{Nationella Språkbanken} -- jointly funded by its 10 partner institutions and the Swedish Research Council (dnr 2017-00626), as well as partly supported by a grant from the Swedish Riksbankens Jubileumsfond (SweLL - research infrastructure for Swedish as a second language, dnr IN16-0464:1).








\bibliographystyle{acl_natbib}
\bibliography{nlp4call2021}

\clearpage
\onecolumn
\section*{Appendices}
\subsection*{Appendix A. Overview of all correction types in the source corpus}
\label{appendix:a}

\begin{figure*}[!htb]
    \centering
    \includegraphics[width=\textwidth]{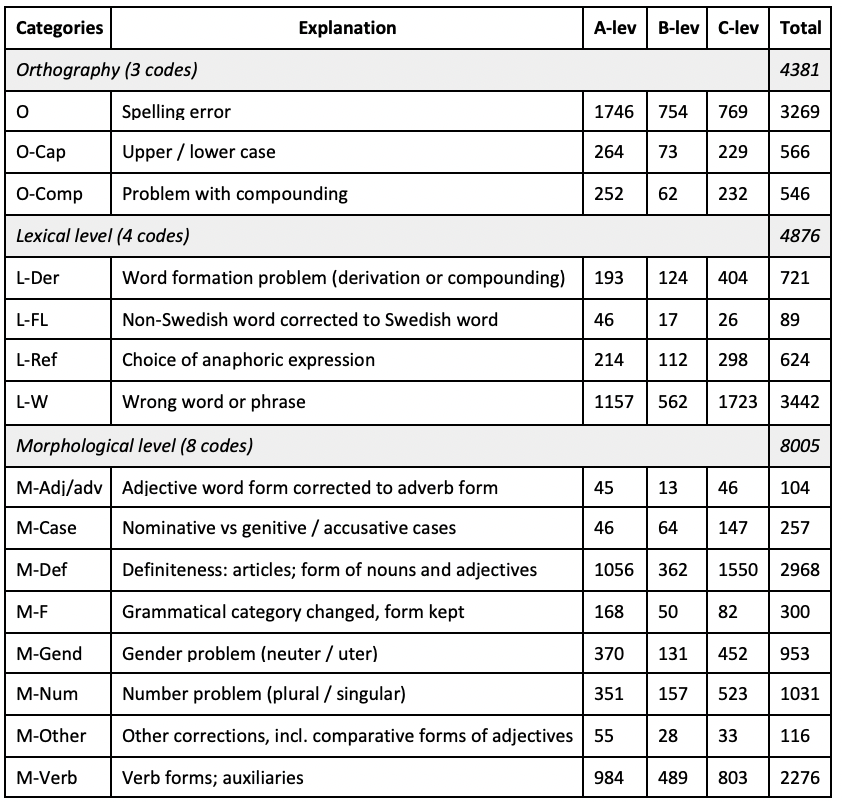}
    \caption{Overview of all correction types in the SweLL corpus, part 1}
    \label{fig:swell-all-categories1}
\end{figure*}

\begin{figure*}
    \centering
    \includegraphics[width=\textwidth]{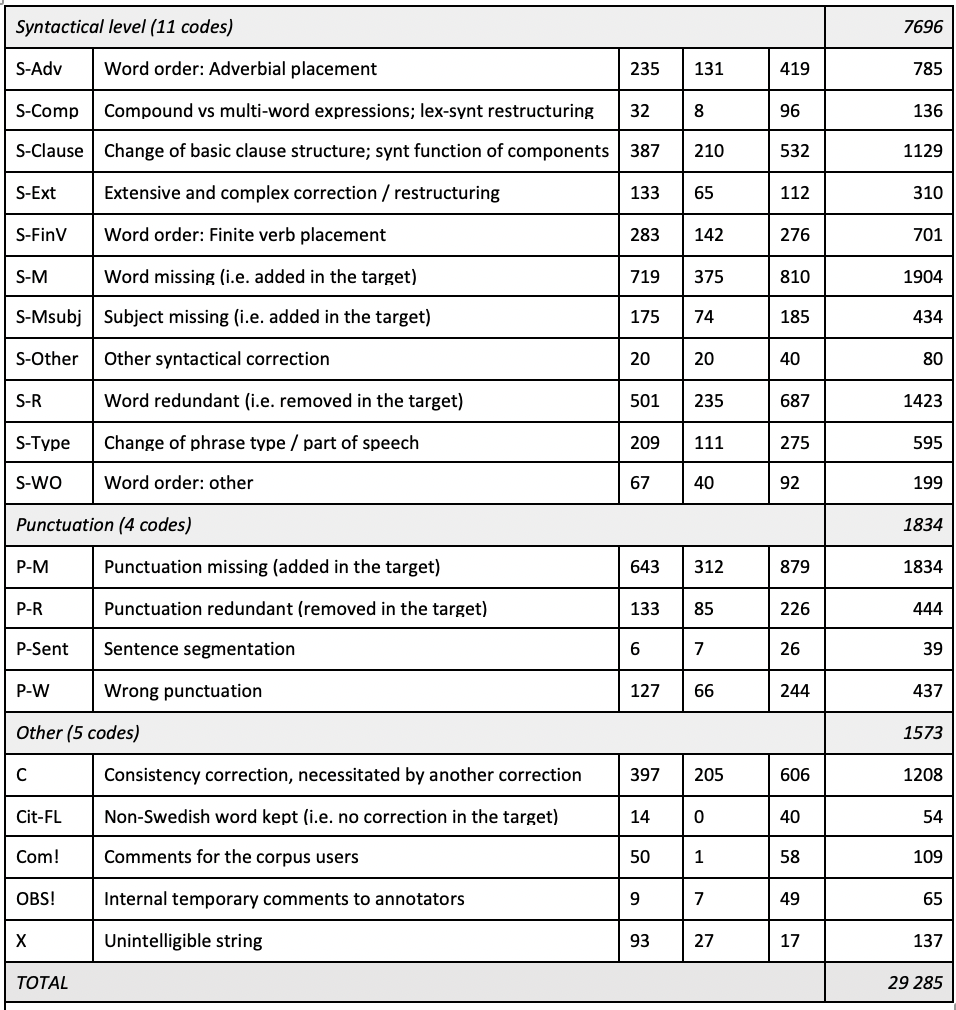}
    \caption{Overview of all correction types in the SweLL corpus, part 2}
    \vspace{128in}
    \label{fig:swell-all-categories2}
\end{figure*}

\end{document}